\documentclass{article}
\usepackage{arxiv}
\usepackage[utf8]{inputenc} % allow utf-8 input
\usepackage[T1]{fontenc}    % use 8-bit T1 fonts
\usepackage{cite}
\usepackage{amsmath,amssymb,amsfonts}
\usepackage{tcolorbox}
\usepackage{algorithmic}
\usepackage{graphicx}
\usepackage{textcomp}
\usepackage{xcolor}
\usepackage{array}
\usepackage{float}
\usepackage{url}
\usepackage{fancyvrb}
\usepackage{booktabs}
\usepackage{mathtools}
\usepackage{doi}
\usepackage{nicefrac}       % compact symbols for 1/2, etc.
\usepackage{microtype}      % microtypography
\makeatletter

\newif\ifieeetemplate
\ieeetemplatefalse

\usepackage{hyphenat}
\graphicspath{ {./figures/} }

\usepackage{hyperref}
\hypersetup{
    colorlinks=false,
    pdftitle={On the Assessment of Benchmark Suites for Algorithm Comparison},
    pdfauthor={D.I. Mattos, L. Ruud, J. Bosch and H.H. Olsson}
}

\title{On the Assessment of Benchmark Suites for Algorithm Comparison}

\author{David Issa Mattos \thanks{\texttt{davidis@chalmers.se} Department of Computer Science and Engineering at Chalmers University of Technology, Gothenburg, Sweden}
\And
Lucas Ruud \thanks{\texttt{lucasr@student.chalmers.se} Department of Computer Science and Engineering at Chalmers University of Technology, Gothenburg, Sweden}
\And
Jan Bosch \thanks{\texttt{jan.bosch@chalmers.se} Department of Computer Science and Engineering at Chalmers University of Technology, Gothenburg, Sweden}
\And
Helena Holmström Olsson
\thanks{\texttt{helena.holmstrom.olsson@mau.se} Department of Computer Science and Media Technology at Malmö University, Malmö, Sweden}
}

\renewcommand{\headeright}{}
\renewcommand{\undertitle}{}
\renewcommand{\shorttitle}{}

\begin{document}
\maketitle

\begin{abstract}
	Benchmark suites, i.e. a collection of benchmark functions, are widely used in the comparison of black-box optimization algorithms. Over the years, research has identified many desired qualities for benchmark suites, such as diverse topology, different difficulties, scalability, representativeness of real-world problems among others. However, while the topology characteristics have been subjected to previous study, there is no study that has statistically evaluated the difficulty level of benchmark functions, how well they discriminate optimization algorithms and how suitable is a benchmark suite for algorithm comparison. In this paper, we propose the use of an item response theory (IRT) model, the Bayesian two-parameter logistic model for multiple attempts, to statistically evaluate these aspects with respect to the empirical success rate of algorithms. With this model, we can assess the difficulty level of each benchmark, how well they discriminate different algorithms, the ability score of an algorithm, and how much information the benchmark suite adds in the estimation of the ability scores. We demonstrate the use of this model in two well-known benchmark suites, the Black-Box Optimization Benchmark (BBOB) for continuous optimization and the Pseudo Boolean Optimization (PBO) for discrete optimization. We found that most benchmark functions of BBOB suite have high difficulty levels (compared to the optimization algorithms) and low discrimination. For the PBO, most functions have good discrimination parameters but are often considered too easy. We discuss potential uses of IRT in benchmarking, including its use to improve the design of benchmark suites, to measure multiple aspects of the algorithms, and to design adaptive suites.
\end{abstract}

% keywords can be removed

\section{Introduction}
\label{sec:introduction}

\ifieeetemplate
    \IEEEPARstart{B}{enchmarking}
\else
    Benchmarking
\fi
studies, and consequently the use of a collection of benchmark functions, have a diverse set of goals, such as visualization and assessment (performance and search behavior, algorithm comparison and competition), analysis of the sensitivity and performance of algorithms (e.g. algorithm tuning), performance extrapolation (e.g. performance regression and automated algorithm design) theory studies (e.g. cross-validation of theoretical studies) and algorithm development (e.g. for code validation) \cite{bartz2020benchmarking}.

The design of a benchmark suite involves many decisions and many properties that affect the usefulness of the suite for different uses. Among them, Bartz-Beielstein \cite{bartz2020benchmarking} cites four core properties. A benchmark suite should be: (1) diverse in terms of characteristics and difficulty for search; (2) representative in respect to real-world problems under investigation; (3) Scalable and tunable; (4) have known solutions or at least information of best performance when the optimal in unknown.

Research has qualitatively evaluated benchmark suites in respect to the properties above \cite{hansen2009real,wang2020iohanalyzer,sala2020benchmarking, munoz2015algorithm,mersmann2010benchmarking, garden2014analysis}. However, in the specific problem of algorithm comparison and competition (i.e. selecting an algorithm with the best performance), there is no research that has statistically evaluated the difficulty level of the benchmark functions in a benchmark suite and how suitable they are for comparing and discriminating algorithms.

This paper addresses this specific gap utilizing item response theory (IRT) statistical models. IRT is an integrated and widely used approach in education and psychology research to develop tests composed of many items to evaluate specific qualities of test-takers \cite{van2016handbook}. In this paper, we provide three main contributions. First, we present an introduction to item response theory. Second, we provide two case studies of the use of IRT to assess commonly used benchmark suites, the BBOB \cite{hansen2009real} and the PBO \cite{doerr2020benchmarking}. In these two cases, we analyze the suitability of the benchmark functions in terms of their difficulty level, discrimination, item information, and test information. Finally, we present a discussion of the use of IRT to help in the selection of benchmark functions and discuss potential extensions and use of IRT in evolutionary computing.

The remainder of this paper is organized as follows. Section \ref{sec:relatedwork} discusses related work. Section \ref{sec:irt} provides an overview of item response theory and Bayesian IRT models. Section \ref{sec:cases} examines two benchmark suites, the BBOB 2019 and the PBO, in the context of IRT. Section \ref{sec:discussion} provides a discussion with potential uses and possible extensions of IRT models that can be used to design benchmark suites. Section \ref{sec:conclusion} concludes this paper. All the data, models, and code used to generate the figures in this paper and other additional content are available at the online appendix at: \url{https://doi.org/10.5281/zenodo.4680443}.

\section{Related work}
\label{sec:relatedwork}

Bartz-Beielstein et al. \cite{bartz2020benchmarking} provide an extensive survey on best practices for benchmarking in optimization. The work covers the different goals, discussion on the choice and desired qualities of benchmark functions and suites, families of algorithms, and how to conduct the analysis of the results. However, from the analysis perspective, it focuses solely on null hypothesis testing and considerations beyond the statistical significance on comparisons, separation of the algorithm performance from the benchmark, and the use of Bayesian statistics for the analysis of the results are not covered. 

Sala and Müller \cite{sala2020benchmarking} discuss the representativeness of the use of artificial benchmark functions to technologically relevant real-world optimization problems since from the theoretical perspective these benchmarks have little generalization value. The work also points as an open challenge considerations about how specific and general classes of benchmark problems should be defined to be of practical use.

Muñoz et al. \cite{munoz2015algorithm} provide an overview of the different topological aspects of  benchmark functions that can affect their difficulty in being solved or time to be solved. Some characteristics are modality, smoothness, presence of plateaus, and the general global structure. Knowledge of these characteristics can help to select algorithms that can lead to better performance. However, in practice, such information might not be available.

Mersmann et al. \cite{mersmann2010benchmarking} provide an exploratory topological analysis of the benchmark functions from the BBOB benchmark suite. The authors classify the functions in terms of modality, global structure, separability, variable scale, search space homogeneity, basin size, global to local optima contrast, and presence of plateaus. These aspects largely influence the probability of solving a problem as well as the expected runtime of an algorithm.

Garden and Engelbrecht \cite{garden2014analysis} utilize  a self-organizing feature map to cluster and analyze benchmark functions of two suites, the BBOB and CEC, in respect to topological properties. The paper finds that while these suites cover a wide range of characteristics,  many characteristics are underrepresented or not even covered. Additionally, both suites are composed of functions that are highly similar to these measured characteristics.

Mattos et al. \cite{mattos2020statistical} discussed the use of Bayesian statistics for data analysis (BDA) of benchmarking in evolutionary computing. They provide an overview of basic concepts on BDA to ensure validity and transparency of the analysis. In this work, we utilize BDA for our item response theory models. We refer to the paper previously mentioned paper and other works in BDA \cite{gelman2013bayesian,mcelreath2020statistical} for concepts such as credible intervals, model checking, and convergence.

Item Response Theory has also  been proposed for the analysis of datasets in machine learning classification problems \cite{martinez2019item}. However, instead of analyzing the entire difficulty or discrimination parameters of the dataset they investigate each individual classification task of each dataset. They utilize a three-parameter logistic model which includes a guessing (or luck) parameter for each item. This extra parameter represents the probability of an algorithm correctly classify an instance even if the difficulty of that instance is much higher than the algorithm's ability. The estimation method is conducted with the maximum likelihood estimator.

\section{Item Response Theory}
\label{sec:irt}

In this section, we provide a short overview of the basics of item response theory. At the end of this section, we compare IRT with the classical test theory.

Item response theory (IRT) was initially developed in the field of educational research to evaluate how latent traits of students (such as intelligence) can be evaluated by a set of items (an exam). The foundations of item response theory come from the idea of utilizing latent variables in education research from Binet (1905) and Thurstone (1925) and were further developed by Lord (1952) and Rasch (1960) \cite{van2016handbook}. Since Lord and Rasch, item response theory has been used in wide range of applications such as large-scale group-score assessment \cite{mazzeo2017large}, psychological testing \cite{de2017psychological}, marketing research \cite{jong2017marketing}, machine learning \cite{martinez2019item} among others.

While there is a large number of IRT models available that addresses a number of problems and applications, we focus on this section on the development and interpretation of the two-parameter logistic model (2PL).

Before introducing the 2PL model, we provide a translation notation of IRT in education and benchmarks in evolutionary computing (BEC). An item ($i$) corresponds to each question in a test. In BEC, an item corresponds to a benchmark function. A test corresponds to a collection of items. In BEC, it corresponds to the benchmark suite. A test taker ($p$) is the subject responding to the test. In BEC, it corresponds to each optimization algorithm. This section utilizes the IRT notation (item, test-taker, and test), while the specific case studies from section \ref{sec:cases} utilizes the evolutionary computing terms.

\subsection{The two-parameter logistic model (2PL)}
The two-parameter logistic model (2PL)  was introduced by Birnbaum (1968) \cite{birnbaum1968some} to model students' abilities when taking an exam. The model assumes that each item has a dichotomous  response, correct or wrong and that exists a latent trait variable (the ability parameter $\theta_p$) that will influence the probability of a test taker ($p$) to correctly answer an item ($i$) over a logistic regression curve \cite{van2016handbook}.

The 2PL model is represented by the equations below \cite{van2016unidimensional}:

\begin{subequations}
\label{eq:2PL}
\begin{align}
    \pi_{p,i} &= \dfrac{\exp (a_i(\theta_p -b_i))}{1+\exp(a_i(\theta_p -b_i))} \\
    y_{p,i} &\sim \text{Bernoulli}(\pi_{p,i})
\end{align}
\end{subequations}

In this model, we have the following notation:
\begin{itemize}
    \item $\pi_{p,i}$ is the probability of an item $i$ being correctly answered by test taker $p$.
    \item $y_{p,i}$ is the dichotomous response from test taker $p$ on item $i$. The value of 1 is for a correct answer and 0 for a wrong answer.
    \item $ a_i$ is the discrimination parameter of item $i$
    \item $b_i$ is the difficulty level of item $i$
    \item $\theta_p$ is the latent trait of the test taker $p$.
\end{itemize}

The 2PL model considers that each item can be assessed in terms of their difficulty level ($b_i$) and on their discrimination level ($a_i$). Below we discuss the impact of each parameter in the probability of a test taker correctly answering an item.

\subsubsection{Item Characteristic Curve}
The probability of model \ref{eq:2PL} plotted over the ability levels of the test takers is called the Item Characteristic Curve. This curve represents the probability of a test taker of a specific ability ($\theta_p$) to correctly answer an item ($i$). This curve is a logistic regression curve controlled by two parameters, the difficulty level (($b_i$))  and the discrimination level ($a_i$).

\begin{figure*}[t!]
\centering
\begin{minipage}{.45\textwidth}
  \centering
  \includegraphics[width=3in]{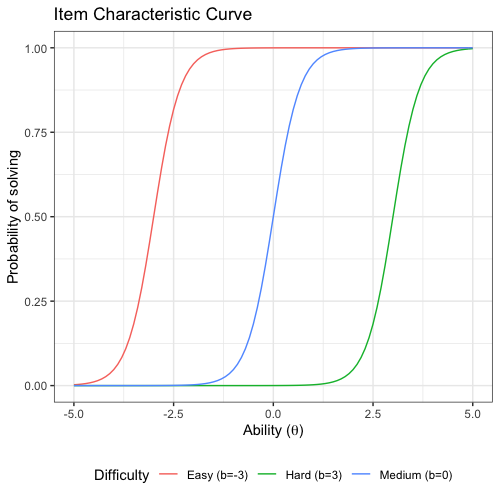}
\end{minipage}%
\begin{minipage}{.45\textwidth}
  \centering
  \includegraphics[width=3in]{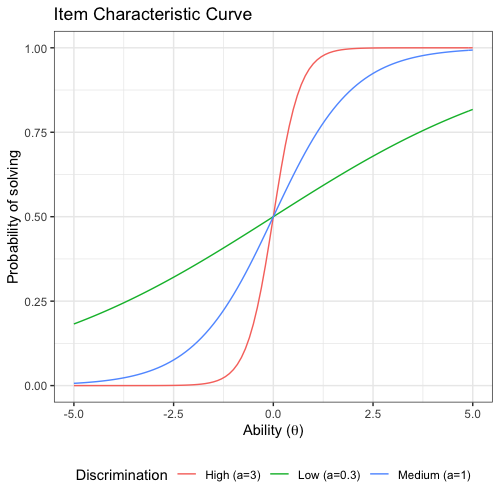}
\end{minipage}
 \caption{Left plot: Impact of the difficulty level in the item characteristic curve. Items with lower difficulty level are shifted to the left and items with higher difficulty level are shifted to the right. The discrimination parameter is $a=3$ for all three curves. Right plot: Impact of the discrimination level in the item characteristic curve. Higher discrimination values require smaller differences in the ability level to differentiate two test takers that have an ability close to the difficulty level of the item. The difficulty level parameter is $b=0$ for all three curves.}
 \label{fig:2pl_discrimination_difficulty}
\end{figure*}

\paragraph{Difficulty parameter}
The difficulty level shifts the logistic curve either to the left (easier items to solve) or right (hard items to solve). Easier items have a higher probability to be correctly answered regardless of the difficulty level. Hard items require a much higher ability level of the respondent $p$ to correctly answer the question. The left plot of Figure \ref{fig:2pl_discrimination_difficulty} shows the impact of the difficulty parameter in the item characteristic curve. It is worth mentioning that the concepts of easy, medium, and hard presented in the figure are relative to the ability parameter $\theta_p=0$.

\paragraph{Discrimination parameter}
The discrimination coefficient indicates the maximum slope of the logistic curve. A higher slope allows a shift in the probability of correctly differentiating the respondents when their ability matches the item's difficulty. The right plot of figure \ref{fig:2pl_discrimination_difficulty} shows the impact of the discrimination parameter in the item characteristic curve.

\subsubsection{Item and test information function}
In IRT, we can utilize the Fisher information \cite{ly2017tutorial} to measure the amount of information each item estimate carries in respect to the ability parameter. This allows us to design test items that maximize the obtained information from the items and better assess the latent trait that differentiates the test takers.

The item information function for dichotomous responses is represented by equation \ref{eq:information}, where $P_i(\theta)$ corresponds to the item $i$ characteristic function \cite{samajima1994estimation}.

\begin{subequations}
\label{eq:information}
\begin{align}
  I_i(\theta) &= \dfrac{[\dfrac{\partial}{\partial \theta} P_i(\theta)]^2}{P_i(\theta)Q_i(\theta)}\\
    Q_i(\theta) &= 1-P_i(\theta)
\end{align}
\end{subequations}

For the two-parameter logistic model, this equation reduces to \cite{de2013theory}:

\begin{equation}
\label{eq:information_2pl}
I_i(\theta) = a_i^2 P_i(\theta)Q_i(\theta)
\end{equation}

Equation \ref{eq:information_2pl} indicates that the item information has a maximum peak at the difficulty level parameter and it is proportional to the square of the discrimination parameter. Higher discrimination increases the information at the difficulty level however it also has a steeper decay.

The test information function ($I(\theta)$) is the conditional expectation of each item information function ($I_i(\theta)$), which when assuming local independence by the items is given by \cite{samajima1994estimation}:

\begin{equation}
\label{eq:test_information_2pl}
I(\theta) = \sum _i I_i(\theta)
\end{equation}

Given that the test information function is the sum of the item information functions, we can design the test based on the individual contributions of each item. By selecting a range of different difficulty levels we can increase specific parts of the information curve so we can better assess the latent variable at that level.

\begin{figure*}[t]
\centering
\begin{minipage}{.45\textwidth}
  \centering
  \includegraphics[width=3in]{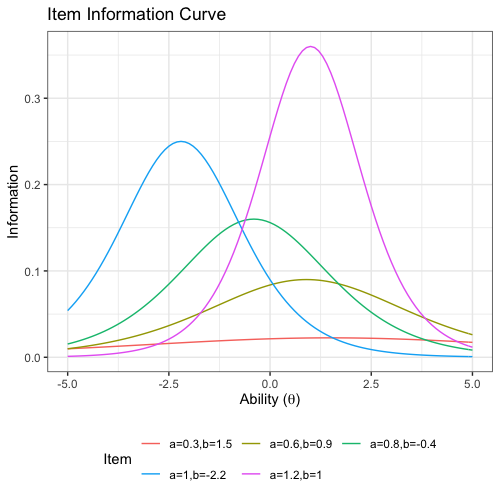}
\end{minipage}%
\begin{minipage}{.45\textwidth}
  \centering
  \includegraphics[width=3in]{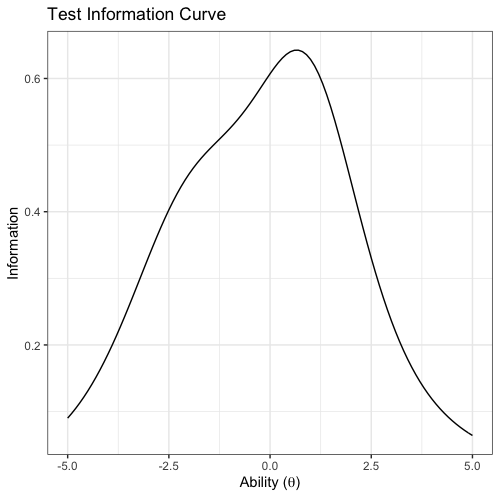}
\end{minipage}
 \caption{The plot on the left shows the item information curve for different difficulty and discrimination parameters. The plot on right shows how these items add to make the test information function. The peak of the test information curve shows the ability regions where the test better estimate test-takers' ability.}
 \label{fig:2pl_info}
\end{figure*}

Figure \ref{fig:2pl_info} shows the information curve for different values of difficulty level and discrimination using the formula from equation \ref{eq:information} as well as the test information curve given a test constructed with the same items.

The test information function in figure \ref{fig:2pl_info} shows that this test is more appropriated to measure test-takers with abilities close to 1 (the peak of the information curve).

The peaks of the test information function indicate the latent trait abilities regions where the test is most able to estimate this trait, i.e. with lower standard errors of measurement (SEM). 

\begin{equation}
\label{eq:sem}
    \text{SEM}(\theta) = \dfrac{1}{\sqrt{I(\theta)}} 
\end{equation}

In the design of a test, we take into account the test information function to select the ability regions and a number of items (which impacts the time taken to conduct the test) where we want to better estimate the latent trait and maximize the information of that region.

\subsection{The Bayesian two-parameter logistic model}
In this paper, we utilize a Bayesian formulation of the 2PL model represented by equations \ref{eq:bayes_2pl}. 

\begin{subequations}
\label{eq:bayes_2pl}
\begin{align}
 y_{i,p} &\sim \text{Binomial}(N_{i,p},\mu_{i,p}) \\
\mu_{i,p} &= \dfrac{\exp(a_i*(b_i-\theta_p))}{1+ \exp(a_i*(b_i-\theta_p))} \\
a_i &\sim \text{Half-}\mathcal{N}(0,5) \\
b_i &\sim \mathcal{N}(0,5)\\
\theta_p &\sim \mathcal{N}(0,5)
\end{align}
\end{subequations}

In this model, we restrict the discrimination parameter to be positive and thus indicating difficulty (as opposed to easiness), and all priors were set to be weakly informative priors \cite{mattos2020statistical}. The Bayesian 2PL model is coded using the Stan probabilistic programming language \cite{carpenter2017stan} and the \texttt{cmdstanr} R package\footnote{\url{https://github.com/stan-dev/cmdstanr}}. To sample the posterior distribution of the parameter of the model, Stan utilizes the No-U-Turn Hamiltonian Monte Carlo sampler \cite{hoffman2014no}. Convergence checks such as traceplots and the Gelman-Rubin potential scale reduction \cite{gelman1992inference} are provided in the online appendix.

\subsection{Classical Test Theory}
Classical test theory is an alternative modeling approach to evaluate tests. At its essence, classical test theory asserts that the observed score is determined by the actual state of a true but unobserved variable plus the random error errors from all other influences \cite{devellis2006classical}. For instance, we can assume that each algorithm has a true probability to solve a specific benchmark problem. When observing the data from benchmark experiments, we obtain the observed probability plus errors. In classical test theory, while we can analyze properties of each item (a benchmark function), its primary focus is on the analysis of the set of items (a suite of benchmark functions) and generating tools to assess the suitability of test, such as reliability indexes.

Classical test theory has two main disadvantages over item response theory. First, that it systematically confounds what we measure (the performance of the algorithms) with the test items used to measure it (the benchmark functions) since it does not create a separation of the parameters for the items and for abilities \cite{van2016handbook}. Second, it aims to create a true score for the entire test \textit{a priori} without taking into account that the test takers (the algorithms) can have different ability levels. 
\section{Benchmarking case studies}
\label{sec:cases}

In this section, we analyze two case studies from well-documented benchmark suites that have archived data, the Black-Box Optimization Benchmarking (BBOB) \cite{hansen2009real} and the Pseudo Boleean Optimization (PBO) benchmark \cite{doerr2020benchmarking}.

For both benchmark suites, we assume the presence of a latent  trait that represents the ability to have empirical success given a fixed budget. Empirical success given a total number of runs is defined as the number of runs in which an algorithm reached the given target \cite{wang2020iohanalyzer}. 

We model the empirical success latent trait of an algorithm as well as the difficulty and discrimination parameters of the benchmark functions utilizing the Bayesian two-parameter logistic model described by equation \ref{eq:bayes_2pl}.

We reinforce that our analysis does not address qualitative and topological issues from the utilization of these benchmark suites but focuses on assessing the suitability of these suites for a general comparison (for the whole suite) of optimization algorithms in terms of the ability to solve a problem. In this section, we present only the plots with the parameter estimates and the test information curve. The parameter tables, item information curve, and validity checks on the convergence of the model are presented in the online appendix. 

\subsection{The BBOB 2019}
\label{sec:bbob}

The Black-Box Optimization Benchmarking (BBOB) workshop series\footnote{http://numbbo.github.io/workshops/index.html} has been held for ten editions. The workshop aims at discussing the last achievements in black-box optimization benchmarking and sharing numerical benchmarking results for continuous and mixed-integer domains. The workshop series is extensively based on the Comparing Continuous Optimization (COCO) platform \cite{hansen2021coco}. The COCO platform provides a consistent platform for running benchmarking experiments in a number of manually designed benchmark suites. The platform also provides a number of archive data of comparative benchmarking data. In this case study, we utilize the archive data available for the noise-free real-parameter single-objective benchmark functions with the algorithms and data from 2019. This data can be accessed at \url{https://numbbo.github.io/data-archive/bbob/}.

\subsubsection{The benchmark suite}
The BBOB suite contains 24 noise-free real-parameter single-objective benchmark functions \cite{hansen2009real}. These functions were designed in mind to evaluate the performance of algorithms with regard to typical difficulties that the authors believe to occur in the continuous domain search. The functions are scalable and designed so algorithms' behavior can be understood in the topological context of the function. All 24 functions are scalable, can be evaluated over $\mathcal{R}^D$ with a search domain between $[-5,5]^D$.

The archive contains data for dimensions 2, 3, 5, 10, 20, and 40. To simplify this analysis we utilize only data where the functions have 5 dimensions.

In this dataset, the algorithms have a maximum budget of 1,000,000 function evaluations per dimension.

\subsubsection{The algorithms}

The selected algorithms with available benchmarking data from 2019 are listed below.

\begin{itemize}
    \item Truncated Newton \cite{varelas2019benchmarking}.
    \item Powell \cite{varelas2019benchmarking}.
    \item Nelder-Mead \cite{varelas2019benchmarking}.
    \item L-BFGS-B \cite{varelas2019benchmarking}.
    \item DE \cite{varelas2019benchmarking}.
    \item Conjugated Gradient (CG) \cite{varelas2019benchmarking}.
    \item COBYLA \cite{varelas2019benchmarking}.
    \item BFGS \cite{varelas2019benchmarking}.
    \item Adaptive Nelder-Mead \cite{varelas2019benchmarking}.
    \item Adaptive Two Mode \cite{bodner2019benchmarking}.
    \item RS-4 \cite{brockhoff2019impact}.
    \item RS-5 \cite{brockhoff2019impact}.
    \item RS-6 \cite{brockhoff2019impact}.
\end{itemize}

\subsubsection{Analysis}
The data was processed utilizing the \texttt{IOHanalyzer} R package \cite{wang2020iohanalyzer}. We utilize the model from equations \ref{eq:bayes_2pl} to estimate the parameters of the model. The model was sampled with 4 chains, 5000 samples, and a warmup of 1000 samples.

\begin{figure*}
\centering
\begin{minipage}{.45\textwidth}
  \centering
  \includegraphics[width=3in]{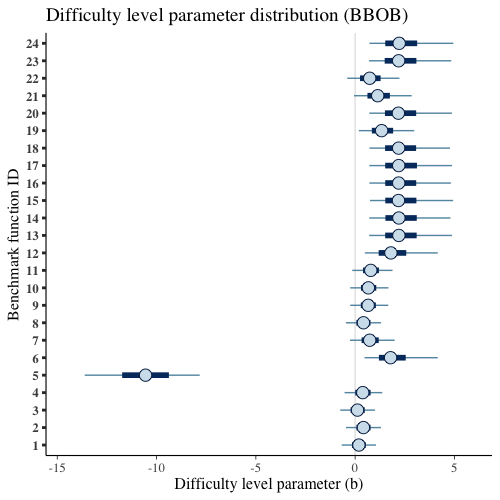}
\end{minipage}%
\begin{minipage}{.45\textwidth}
  \centering
  \includegraphics[width=3in]{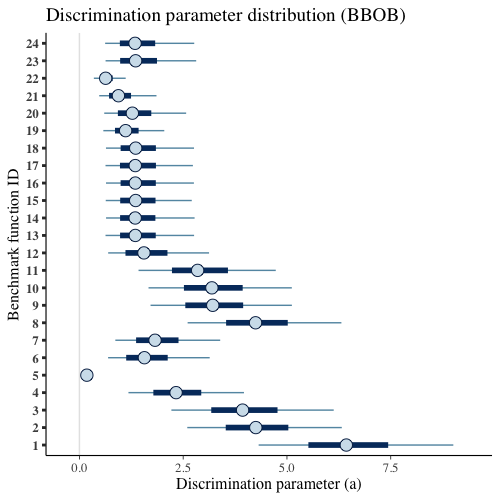}
\end{minipage}
 \caption{The plot on the left shows the difficulty level and the plot on the right shows the discrimination parameter of the BBOB benchmark functions. The thin blue line corresponds to the 90\% credible interval and the thick blue line corresponds to the 50\% credible interval}
 \label{fig:bbob_diff_disc}
\end{figure*}

\paragraph{Difficulty and discrimination}
The posterior distribution of the discrimination and difficulty parameters for each of the 24 benchmark functions are estimated and shown in figure \ref{fig:bbob_diff_disc}. From this figure, we can see that most functions have a positive difficulty parameter between 0 and 5. Functions 1, 2, 3, 4, 22 have difficulty levels close to zero between 0 and 5. Only function 5 has a negative difficulty parameter. In terms of discrimination, most functions have a low discrimination parameter with a median close to 1. 

\begin{figure}
    \centering
    \includegraphics[width=3in]{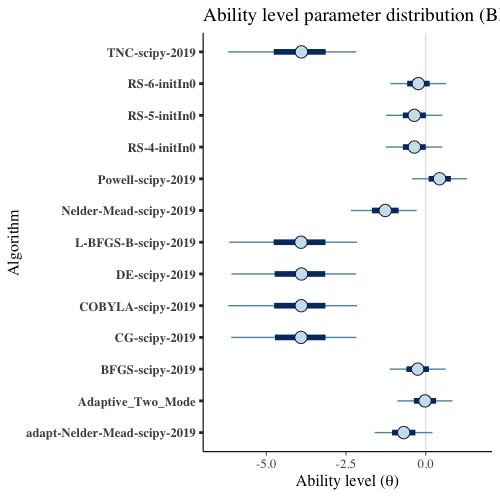}
    \caption{Estimated ability of the algorithms included in the BBOB 2019 benchmark dataset. The thin blue line corresponds to the 90\% credible interval and the thick blue line corresponds to the 50\% credible interval}
    \label{fig:bbob_abil}
\end{figure}

\paragraph{The ability of the algorithms}
The posterior distribution of the ability parameter of the algorithms are estimated and shown in figure \ref{fig:bbob_abil}. We can see that most algorithms have a negative ability or close to zero. The large overlapping intervals between some groups of algorithms indicate that they do not have a significantly different ability level. The algorithm with the highest ability to solve a problem is the Powell algorithm, which has a median ability of 0.3.

\begin{figure*}
    \centering
    \includegraphics[width=0.8\textwidth]{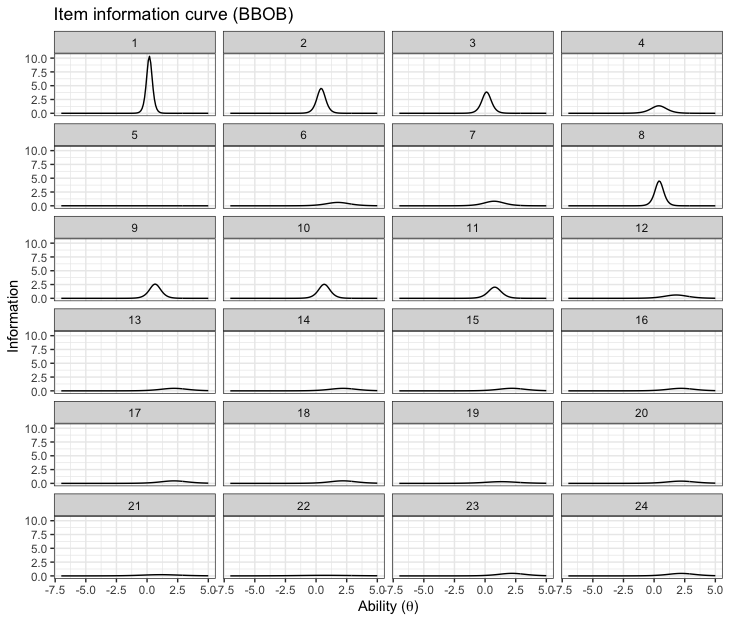}
    \caption{Median item information curve for all BBOB benchmark functions. This figure indicates the presence of many benchmark functions that add little information to the capability of the test to estimate the empirical success ability of the algorithms.}
    \label{fig:bbob_item_info}
\end{figure*}

\paragraph{Item information curves}
Based on the obtained parameters we can create the item information curves utilizing equation \ref{eq:information_2pl}. Figure \ref{fig:bbob_item_info} shows the item information curve for all 24 benchmark functions.

\begin{figure*}
    \centering
    \includegraphics[width=0.8\textwidth]{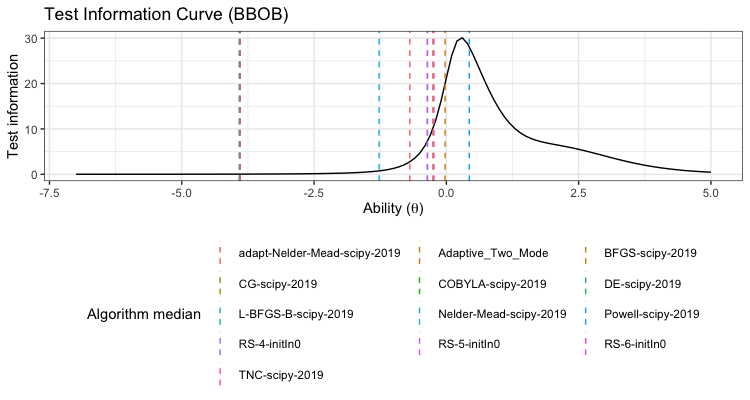}
    \caption{Test information curve (with the median parameters). The median ability of the algorithms is represented by the dashed vertical lines. This figure indicates that many algorithms are in a region were the test does not estimate well the empirical success ability of the algorithms. Additionally, the available information comes only a few benchmark functions.}
    \label{fig:bbob_test_info}
\end{figure*}

\paragraph{Test information}
Utilizing equation \ref{eq:test_information_2pl} and the item information curves we can obtain the test information curve. Figure \ref{fig:bbob_test_info} shows the median test information curve together with the median abilities of the algorithms. 

\subsubsection{Discussion}

The low value of the discrimination parameters is reflected by the high difficulty parameter of these functions. If a particular item has a very high difficulty level or a very low difficulty level compared to the abilities of the algorithms, the response of the algorithms will be very similar and therefore have a low estimation of the discrimination parameters. 

The test information curve shows that this benchmark suite is more suitable for algorithms with an ability close to zero. Given that most of the algorithms have a negative median ability this benchmark suite is not appropriated to compare the probability of solving a problem for these algorithms. By looking at the individual item information curve (available in the online appendix), most of the test information comes from functions 1, 2, 3, 8, 9, 10, and 11, which are the functions with higher discrimination parameters and difficulty level closer to the abilities of the algorithm. The other functions add little information to the estimates of the empirical probability of success.

In the appendix, we also provide additional analysis of the BBOB suite utilizing the data archive from 2009 and 2019. This additional analysis was conducted to observe the impact of the selection of algorithms since the BBOB 2009 archive contains a larger number of algorithms (30). Nevertheless, the results obtained there are consistent with the BBOB 2019 analysis we provide and discuss here. 

\subsection{The PBO}
\label{sec:pbo}
The Pseudo-Boolean Optimization benchmark suite is a collection of benchmark functions for black-box discrete optimization that is part of the IOHprofiler benchmarking environment \cite{doerr2020benchmarking}. The functions designed in the IOHprofiler platform were designed to discriminate the performance of different black-box discrete optimization algorithms using pseudo-Boolean functions. The goal is to construct a benchmark suite that covers a wide range of the problem characteristics found in real-world combinatorial optimization problems.

The IOHprofiler benchmarking environment also provides a collection of twelve algorithms, that have known strengths and weaknesses, to serve as future reference.

We utilize in this analysis the 11 instances with a single run dataset. This archive data from this benchmark suite can be accessed at \url{https://github.com/IOHprofiler/IOHdata}.

\subsubsection{The benchmark suite}
The PBO benchmark suite contains 23 scalable pseudo-Boolean functions, i.e., the functions are expressed as $f: \{0,1\}^D \mapsto \mathcal{R}$. The authors argue that functions that do not help to do discrimination between the tested algorithms should be evaluated as obsolete although this depends on the collection of algorithms. 

\subsubsection{The algorithms}
We utilize the twelve algorithms proposed and available in the archive data from the PBO benchmark suite \cite{doerr2020benchmarking}.

\begin{itemize}
    \item Greedy Hill Climber (gHC).
    \item Randomized Local Search (RLS).
    \item (1+1) EA.
    \item Fast GA (fGA) with $\beta=1.5$ mutation strength.
    \item (1+10) EA.
    \item (1+10) EA\textsubscript{r/2,2r}.
    \item (1+10) EA\textsubscript{norm}.
    \item (1+10) EA\textsubscript{var}.
    \item (1+10) EA\textsubscript{log-n}.
    \item (1+($\lambda,\lambda$)) GA.
    \item (30,30) vanilla GA (vGA).
    \item Univariate Marginal Distribution Algorithm (UMDA).
\end{itemize}

\subsubsection{Analysis}
The data was processed utilizing the \texttt{IOHanalyzer} R package \cite{wang2020iohanalyzer}. We utilize the model from equations \ref{eq:bayes_2pl} to estimate the parameters of the model. The model was sampled with 4 chains, 5000 samples, and a warmup of 1000 samples.

The archive contains data for dimensions 16, 64, 100, and 625. To simplify this analysis we utilize only data where the functions have 16 dimensions.

\begin{figure*}
\centering
\begin{minipage}{.45\textwidth}
  \centering
  \includegraphics[width=3in]{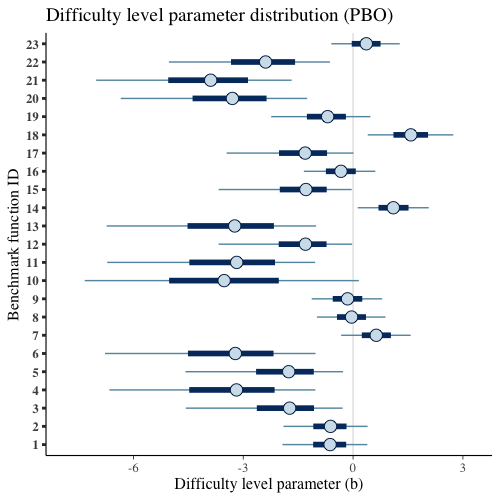}
\end{minipage}%
\begin{minipage}{.45\textwidth}
  \centering
  \includegraphics[width=3in]{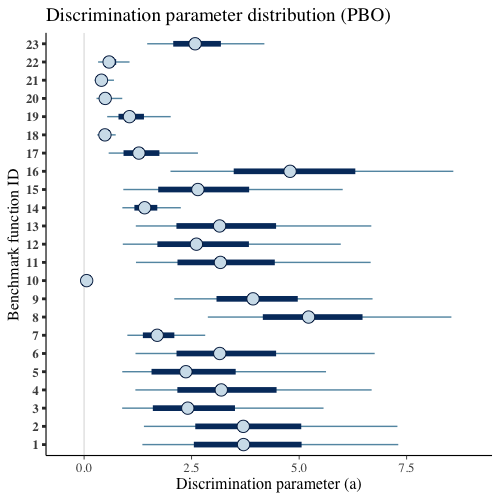}
\end{minipage}
 \caption{The plot on the left shows the difficulty level and the plot on the right shows the discrimination parameter of the PBO benchmark functions. The thin blue line corresponds to the 90\% credible interval and the thick blue line corresponds to the 50\% credible interval}
 \label{fig:pbo_diff_disc}
\end{figure*}

\paragraph{Difficulty and discrimination}
The posterior distribution of the discrimination and difficulty parameters for each of the 23 benchmark functions are estimated and shown in figure \ref{fig:pbo_diff_disc}. From this figure, we can see that most functions have a negative difficulty parameter. Only functions  7, 14, 18, and 23 have more than 50\% of their probability mass above zero. In terms of discrimination, most functions have a high but uncertain, due to the wide intervals, discrimination parameter. 

\paragraph{The ability of the algorithms}

\begin{figure}
    \centering
    \includegraphics[width=3in]{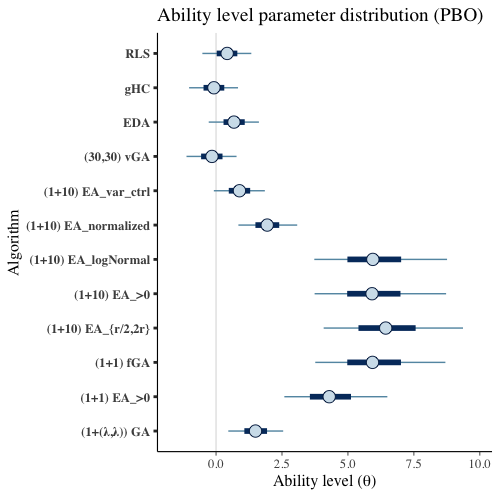}
    \caption{Estimated ability of the algorithms included in the PBO benchmark dataset. The thin blue line corresponds to the 90\% credible interval and the thick blue line corresponds to the 50\% credible interval.}
    \label{fig:pbo_abil}
\end{figure}

The posterior distribution of the ability parameter of the algorithms are estimated and shown in figure \ref{fig:pbo_abil}. We can see that most algorithms have a positive ability level while some algorithms have a high ability level with a 50\% probability mass between 5 and 7.5

\begin{figure*}
    \centering
    \includegraphics[width=0.8\textwidth]{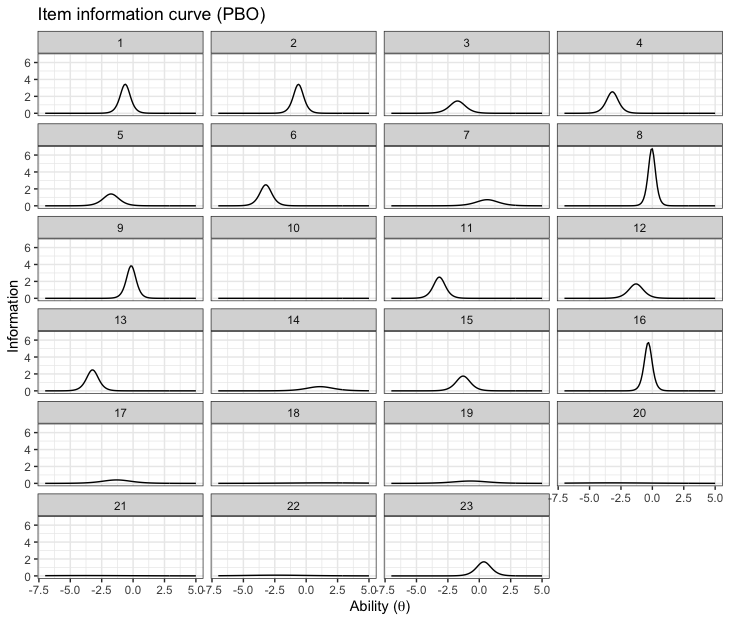}
    \caption{Median item information curve for all PBO benchmark functions. This figure indicates the presence of many functions that add little information to estimating the empirical success ability of the algorithms.}
    \label{fig:pbo_item_info}
\end{figure*}

\paragraph{Item information curves}
Based on the obtained parameters we can create the item information curves utilizing equation \ref{eq:information_2pl}. Figure \ref{fig:pbo_item_info} shows the item information curve for all 23 benchmark functions.

\begin{figure*}
    \centering
    \includegraphics[width=0.8\textwidth]{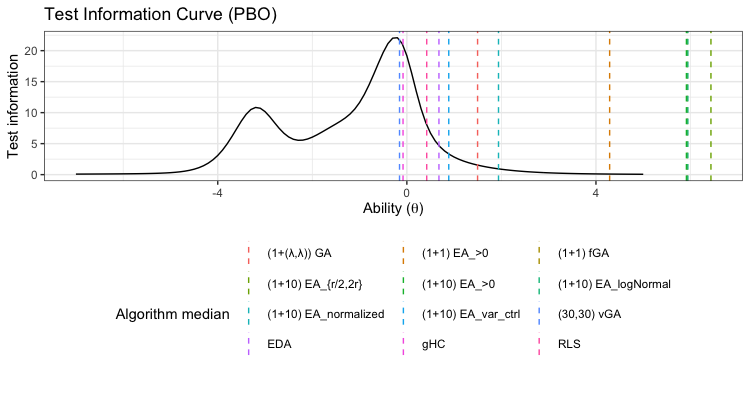}
    \caption{Test information curve (with the median parameters). The median ability of the algorithms is represented by the dashed vertical lines. This figure indicates that most of the algorithms have an ability level far superior to the region were the test has higher measurement accuracy.}
    \label{fig:pbo_test_info}
\end{figure*}

\paragraph{Test information}
Utilizing equation \ref{eq:test_information_2pl} and the item information curves we can obtain the test information curve. Figure \ref{fig:pbo_test_info} shows the median test information curve together with the median abilities of the algorithms for the PBO benchmark functions.

\subsubsection{Discussion}

Most of the algorithms have an ability level much higher than the difficulty level of the proposed benchmark functions. This mismatch results in most algorithms having a very high probability of solving the problems within the suggested budget. This can also be seen from the test information curve, which shows that most of the test information is being measured for algorithms with a negative ability level. However, most of the algorithms have a positive and significantly higher median ability and the test provides zero information for these algorithms. In the appendix, we can observe the item information curve. There we see that functions 7, 10, 14, 17, 18, 19, 20, 21 and 22 provide little information on the estimate of the ability level of the algorithms.

In the online appendix, we show the same analysis for the benchmark functions with 625 dimensions. With the increase in the number of dimensions, the overall difficulty level of the functions increase (they become more negative), and the ability level of the algorithms also reduce. However, most still have an ability far higher than the difficulty level and the test provides little information for most algorithms.
\section{Discussion}
\label{sec:discussion}

In this section, we discuss the use and potential uses, and extensions of item response theory for benchmarking.

\subsection{Use of IRT to design benchmark suites}
Item response theory provides a concise and unified framework to investigate item properties such as difficulty and discrimination, and to help design better tests. By utilizing the item's and test information curves, researchers can evaluate the impact of each benchmark function in the estimation of the latent trait. 

Previous research \cite{munoz2015algorithm,bartz2020benchmarking,mersmann2010benchmarking,garden2014analysis} has mainly focusing on the analysis of the topological properties of individual benchmark functions to indirectly infer difficulty, discrimination and representativeness. The analysis of the benchmark suite has been conducted in terms of qualitative discussion of the presence of those properties. Item response theory, on contrast, provides a quantitative assessment of the suitability of individual benchmark functions in terms of difficulty and discrimination and how each function contributes in terms of information for algorithm comparison in the whole benchmark suite.

In the case studies discussed in the previous section, the test information curve indicates a mismatch between the algorithm's ability level and the benchmark suite. In the case of the BBOB, most benchmarks are considered too hard for the algorithms. In the case of the PBO, the benchmarks are too easy. In both cases, this results in algorithms being evaluated in low information regions. The item information curves presented in the online appendix indicate that multiple benchmark functions add little information to the test and could be replaced by others that can better compare the algorithms in terms of the empirical probability of success. By adjusting the test items, based on a large selection of relevant algorithms (in the ability level where they want to investigate), researchers can design benchmark suites that are more suitable to compare optimization algorithms.

An increase in the number of benchmark functions in a suite will increase the information obtained from the benchmark suite at the expense of time taken to evaluate the benchmark suite. In this context, two decisions can be made while adding or replacing new benchmark functions to a benchmark suite. First, it is to increase information in a specific area so we have fewer measurement errors in that area (creating a higher peak). Second, it is to increase information for additional areas so we can measure better a wider range of algorithms abilities.

\subsection{Latent traits and IRT extensions for benchmarking}
One of the assumptions of the 2PL model used in this work is that it exists a latent ability trait of the algorithms that influence the probability of solving benchmark problems. While this assumption allows us to directly compare algorithms and understand the impact of each item in terms of difficulty and discrimination, it is not the only latent variable of interest. 

The choice of different metrics, for example, the number of function evaluations to converge or improve over random search \cite{mattos2020statistical} among others, can lead to the formulation of multiple latent traits that investigates different aspects of an algorithm. 

However, it is worth noting that while the empirical success rate  has a direct impact in other metrics and performance indicators, such as the Expected Running Time (ERT) \cite{wang2020iohanalyzer}. The ERT takes into account the number of successful runs and can even assume infinity values when none of the runs was successful. Benchmark suites such as the BBOB, which consists of many hard problems for the current level of the algorithms, can significantly increase the measurement error of a latent trait for the ERT. 

Multidimensional Item Response Theory (MIRT) \cite{van2016handbook} is a general class of theories that maps the test responses of a test taker to a constructed space that are the targets of measurement of the test that are the different latent traits. Research on IRT has also provided a number of alternative models that can deal with non-binary responses such as ordered, continuous, and time responses (where the time to complete a test is also of interest) to hierarchical and generalized models.

\subsection{Computerized adaptive tests}
Another novel application of IRT for benchmarking is the creation of computerized adaptive testing (CAT) \cite{linden2000computerized}. With the improvement and consolidation of algorithm traits, adaptive benchmark suites can be designed for the assessment of the latent traits. 

A CAT system would make the selection of the next items in a test based on the information gained and the current estimates of the latent traits based on the previously applied questions. This allows the algorithms to be evaluated in benchmarks that are suitable for their ability level. The latent traits using a CAT are comparable with other latent traits even if the same algorithms have not been evaluated in the same set of benchmark functions. This would allow researchers to quickly assess how their algorithms compare to existing algorithms utilizing a smaller number of benchmark evaluations.

\section{Conclusion}
\label{sec:conclusion}

While previous research has explored and analyzed the topological aspects of the benchmark functions used in common benchmark suites. No work has statistically evaluated how well those functions are suitable for comparing algorithms. In this work, we provide an overview of item response theory modeling and how it can be used to analyze and improve the design of benchmark suites. 

We applied the Bayesian two-parameter logistic model for multiple attempts in datasets from two benchmark suites, the BBOB and the PBO. The analysis of the item information functions and the test information functions of both suites indicate that most of the proposed benchmarks are not suitable for comparing the current state-of-the-art algorithms, since they either have a big mismatch in the difficulty level (too easy for the PBO or too difficult for the BBOB) or have low discrimination. While our analysis does not discuss the adequacy of these benchmark suites for qualitative analysis to understand the behavior of algorithms, we question the suitability of these benchmark suites for algorithm comparison under the empirical success rate metric. 

Item response theory not only helps to evaluate the suitability of the benchmark suites but also provides a systematic and quantitative way to help design appropriate benchmark suites..

\section*{Acknowledgment}
This work was partially supported by the Wallenberg Artificial Intelligence, Autonomous Systems and Software Program (WASP) funded by the Knut and Alice Wallenberg Foundation and by the Software Center.

\bibliographystyle{./bib/IEEEtran}
\bibliography{./bib/bibliography}

\end{document}